# HSI Image Enhancement Classification Based on Knowledge Distillation: A Study on Forgetting

Zhu Songfeng[1]

(1.School of Software Engineering, Henan Polytechnic University, Henan 454000，china)

Abstract

In incremental classification tasks for hyperspectral images, catastrophic forgetting is an unavoidable challenge. While memory recall methods can mitigate this issue, they heavily rely on samples from old categories. This paper proposes a teacher-based knowledge retention method for incremental image classification. It alleviates model forgetting of old category samples by utilizing incremental category samples, without depending on old category samples. Additionally, this paper introduces a mask-based partial category knowledge distillation algorithm. By decoupling knowledge distillation, this approach filters out potentially misleading information that could misguide the student model, thereby enhancing overall accuracy. Comparative and ablation experiments demonstrate the proposed method's robust performance.

1. Introduction

Hyperspectral images (HSI) constitute a data type abundant in spectral information, distinguishing them from conventional natural images that contain merely three bands: red, green, and blue. Hyperspectral images encompass hundreds or even thousands of bands, utilizing not only spatial information but also extensive spectral data. This enables them to attain superior classification capabilities in comparison to natural images [1]. Hyperspectral Image Classification (HSIC) [2][3] constitutes a fundamental technology within the domain of hyperspectral image applications, serving a critical function in areas such as smart agriculture [4], ocean monitoring[5], and mineral exploration[6]. With ongoing scientific and technological advancements, the capabilities for acquiring remote sensing data have progressively enhanced, resulting in the development of new classes. Consequently, traditional HSIC methods are increasingly insufficient to satisfy contemporary application requirements. Given the rapid escalation in the rate of acquiring hyperspectral imaging (HSI) data, the model must integrate samples from new classes into its training process in order to adapt to emerging classes. However, in practical applications, models newly trained often tend to forget information related to previously learned classes, resulting in catastrophic forgetting[7][8]. This issue significantly impairs the effectiveness of HSI classification and impedes its advancement.

To mitigate catastrophic forgetting, the concept of incremental learning (CL) [10]has been introduced. CL consists of two phases: an initial training phase and an incremental learning phase. Its fundamental principle entails maintaining knowledge of previously learned classes while integrating new ones[11][12]. Based on the differences between new and old data, CL can be broadly categorized into three scenarios[13]: task incremental learning, domain incremental learning, and class incremental learning. 1) In the task incremental learning scenario, the focus is on learning entirely new, independent tasks, with each task having distinct boundaries. 2) In the domain incremental learning scenario, the focus is on cross-domain data for the same task. That is, the task objective remains unchanged, but the data distribution is significantly shifted due to factors such as the environment or the device. 3) In the class incremental learning scenario, the goal is to expand the classes of a task. That is, the task objective remains unchanged, but new classes are added, expanding the label space. This paper primarily concentrates on class incremental learning. Existing incremental learning techniques can be broadly classified into three classes[9]: regularization methods, dynamic framework methods, and memory replay methods. 1) Regularization methods impose constraints on the weights of neural networks[14][15][16][17][18]. 2) Dynamic architecture methods offer different model parameters to accommodate various data dynamics[19][20][21][22][23]. 3) Memory recall methods store knowledge of previously learned classes from prior training episodes[24][25][26].

Knowledge Distillation (KD)[27] was initially developed for model compression, with the objective of constructing a more lightweight model through the utilization of supervised information derived from larger models. Subsequently, LWF[16]effectively applied KD to incremental learning for the first time, showcasing the beneficial impact of KD strategies in alleviating the issue of severe forgetting. Distillation schemes can be broadly categorized into two types[28]: offline distillation and online distillation. 1) Offline Distillation: This primarily focuses on improving different aspects of knowledge transfer, such as knowledge design[27] and loss functions for distribution matching[29]. The main advantage of offline distillation lies in its simplicity and ease of implementation, but it suffers from issues like the student model's dependence on the teacher model and lengthy training times. 2) Online Distillation: Concerns regarding offline distillation have drawn attention from the research community[30]. To overcome these limitations, researchers proposed online distillation to further enhance model performance. In online distillation, both teacher and student models are updated synchronously, and the entire knowledge distillation framework is end-to-end trainable. In recent years, numerous online distillation methods have been proposed[31][32][33]. For instance, to enhance generalization capabilities, Guo et al.[34]extended deep mutual learning with soft logits ensembles; to reduce computational costs, Zhu et al.[35]introduced a multi-branch architecture where each branch represents a student model, with different branches sharing the same backbone network.

Throughout research on feature-based knowledge distillation (KD) and logit-based KD, as feature-based KD has exhibited superior performance relative to logit-based KD, the majority of scholarly efforts have concentrated on deriving knowledge from deep features within intermediate layers. However, feature-based methods entail greater training expenses owing to the supplementary computational and storage requirements during the training process.[48]posits that certain factors limit the potential of logit distillation, resulting in suboptimal outcomes. Based on the investigation presented in[48], a decoupled knowledge distillation method has been proposed to overcome the limitations of logit distillation and enhance its effectiveness. This method segregates knowledge distillation into Target Classification Knowledge Distillation (TCKD) and Non-Target Classification Knowledge Distillation (NCKD), thereby removing the constraints between these components to prevent mutual suppression.

Although current incremental learning methods have achieved some advancements in mitigating catastrophic forgetting, numerous obstacles and challenges persist. For example, some datasets are not publicly available owing to privacy considerations or legal constraints, while extended model iterations lead to outdated data, thereby consuming significant storage and computational resources. Consequently, there is a pressing necessity for methods that do not depend on historical datasets. To address concerns related to data privacy, as well as the consumption of storage and computational resources, this paper introduces a Teacher-Based Historical Knowledge Retention (TBHKR) method. Historical knowledge refers to the knowledge acquired by the model subsequent to training on previous class samples. This method adjusts the labels of data points that bear a high similarity to old classes during the incremental learning process. As a result, it effectively simulates training on old class samples from the initial stage without the necessity of involving those samples in the initial training phase. Furthermore, while decoupled knowledge distillation harnesses the capabilities of logit distillation, certain "errors" emerge during the incremental learning process rooted in knowledge distillation. These errors arise because the teacher model delivers outputs not solely for the initial classes but also for the incremental classes throughout the learning process. However, these incremental classes are entirely unfamiliar to the teacher model, making its outputs inherently inaccurate. The inclusion of these erroneous outputs in the distillation loss calculation inevitably introduces bias into the loss computation.

To address the issue of "errors" arising during knowledge distillation-based incremental learning, this paper proposes a decoupled knowledge distillation method based on masking (DKDBM). This method eliminates the influence of erroneous information on the student network, thereby correcting the distillation loss calculation and improving model accuracy. Overall, the contributions of this paper are as follows:

1) A teacher-based knowledge retention method for incremental image classification is proposed. During the incremental learning phase, this method identifies relevant knowledge about the initial class from incremental class samples using a similarity metric based on the teacher network, thereby preserving knowledge about the initial class. This method effectively mitigates catastrophic forgetting while also addressing dependencies on storage resources and data.

2) A partial class knowledge distillation algorithm based on masking is proposed. This method achieves the dual objectives of fully leveraging valuable knowledge learned by the teacher while simultaneously preventing misleading knowledge generated by the teacher from affecting the student model. This is accomplished by filtering out erroneous information during the incremental learning phase.

3) This paper conducted experiments on three widely-used hyperspectral classification datasets, comparing the proposed algorithm with multiple existing methods, while also performing ablation experiments. The experimental results sufficiently demonstrate the effectiveness and efficiency of the proposed algorithm.

2. Related Work

A. Incremental Learning

In recent years, numerous incremental learning methods have been proposed[36], including sample replay-based methods and knowledge distillation-based methods. 1) Sample replay methods maintain a subset of samples from previous classes within a memory buffer during the initial training phase. These samples are subsequently integrated into the training process of new models to prevent the phenomenon of model forgetting old classes[36]. The stored samples typically represent the most characteristic examples of each class. For instance, Rebuffi et al.[24]proposed iCaRL, which selects samples nearest to the class centroid for storage, as these are deemed most representative. Hou et al.[25]identified an imbalance between old and new data as a primary factor contributing to catastrophic forgetting. They proposed cosine normalization, a "few-forget" constraint, and inter-class separation techniques to mitigate this imbalance. Luo et al.[26]introduced CIM, which addresses the limitations in incremental learning where only a limited number of old-class class samples and a small subset of all new-class samples are accessible, by compressing the samples. Nevertheless, this method remains significantly constrained for hyperspectral image classification (HSIC) tasks[37]. First, hyperspectral data exhibits extremely high dimensionality, with the storage of vast quantities of information requiring substantial storage capacity. Furthermore, rapid advancements in remote sensing technology have enabled researchers to continuously acquire large volumes of hyperspectral imagery[38]. Persistent sample replay exacerbates storage demands and markedly increases computational overhead. Second, some hyperspectral images may encompass sensitive geographic or remote sensing information, and their storage and processing could contravene security protocols. Consequently, there exists an urgent necessity for methods that do not rely on historical samples to address these challenges. Li et al.[39]proposed that model inversion techniques could extract information about old class samples from the network without direct access, utilizing this information to generate or reconstruct synthetic images akin to the initial training data. Inspired by this approach, this paper introduces a straightforward yet innovative concept. The proposed method utilizes samples from the incremental learning phase that resemble those from the initial training phase, thereby serving as surrogate samples for the baseline training data.

2) Knowledge distillation-based methods mitigate catastrophic forgetting by enforcing consistency between the outputs of old and new tasks. In this method, the student network is able to emulate the activations of the teacher network during the acquisition of new classes, utilizing soft labels generated by the teacher network, as outlined in LWF[16]. The CSSFD proposed by Zhao et al.[40]employs a spectral distillation strategy aimed at preserving knowledge from the trained model. Furthermore, it incorporates learning metrics into multi-stage feature extraction to reduce discrepancies between old and new models. The DCPRN, proposed by Yu et al.[41]employs a prototype representation mechanism to connect the two phases of incremental learning and introduces Dual Knowledge Distillation (DKD), which synthesizes information from both feature and decision levels. Although various knowledge distillation strategies have been proposed for specific scenarios, the challenge of model forgetting of old classes persists in hyperspectral incremental learning as new classes continue to be incorporated.

B. Knowledge Distillation

Knowledge distillation, as proposed by Hinton et al.[27], is a technique for model compression that can be broadly classified into four types: logit-based knowledge distillation, feature-based knowledge distillation[42], relationship-based knowledge distillation[43], and zero-shot knowledge distillation[44]. 1) Logit-based knowledge distillation allows the student model to emulate the predictions of the teacher model by minimizing the discrepancy between their respective final outputs. 2) Feature-based knowledge distillation emphasizes the alignment of hidden layer features between the student and teacher models. 3) Relational knowledge distillation focuses on transferring the structural relationships between samples or contexts from the teacher model to the student model. 4) Data-free knowledge distillation seeks to transfer knowledge from the teacher to the student without the need for access to initial training samples from the initial training phase. Initially, the initial training data for knowledge distillation was reconstructed using the teacher model and its metadata[45]. However, as metadata is often unavailable, alternative methods have been proposed to optimize noisy images for the generation of synthetic data[46]. Furthermore, methods utilizing generative networks have been introduced[47]; however, these methods still require the separate development of an additional model. Conversely, the straightforward and efficient method presented in this paper accomplishes the goal of removing the reliance on old class samples from the initial training phase, without the need to construct a new, independent model.

In traditional knowledge distillation, the KD loss generally exists in a highly coupled form. However,[48]disaggregates the classical KD loss into two distinct components: Target-Class Knowledge Distillation (TCKD) and Non-Target-Class Knowledge Distillation (NCKD), thereby separating their respective constraints.TCKD conveys knowledge about the "difficulty" of training samples[48], transmitting this knowledge through binary logit distillation-that is, by providing only predictions for the target class while keeping the specific predictions for each non-target class unknown. Conversely, NCKD concentrates exclusively on knowledge contained within non-target logits. The weighting of the NCKD loss term is inversely related to the teacher model's confidence in the target class predictions. Nonetheless, this formulation introduces certain challenges. For example, in NCKD, an increased confidence from the teacher in a sample signifies more reliable and valuable information. However, according to the negative correlation, this scenario would unduly suppress the teacher model. Consequently, adjustments to the weights of both components are warranted.

Since the significance of TCKD and NCKD is coupled, paper [48] assigns distinct weights to both, with their weight values being independent. Specifically, the weight of NCKD is replaced with a constant value to prevent it from being suppressed by TCKD, a technique termed decoupled knowledge distillation. Inspired by decoupled knowledge distillation, this paper aims to apply this decoupling concept to incremental learning by breaking the coupling between the old class samples from the initial training phase and the new class samples from the incremental learning phase. Therefore, we propose a masked partial class knowledge distillation algorithm to filter out erroneous information during incremental learning, preventing the student model from being disturbed by incorrect data. Given the intertwined significance of TCKD and NCKD, paper[48]assigns separate weights to both, with these weights being uncorrelated. Specifically, the weight of NCKD is replaced with a constant value to prevent its suppression by TCKD, a technique known as decoupled knowledge distillation. Inspired by this approach, the present study seeks to extend the concept of decoupling to incremental learning by severing the coupling between old class samples from the initial training phase and new class samples from the incremental learning phase. Accordingly, we propose a masked partial class knowledge distillation algorithm designed to filter out erroneous information during incremental learning, thereby preventing the student model from being adversely affected by incorrect data.

3. Methodology

The problem of catastrophic forgetting constitutes a significant challenge in incremental learning. As hyperspectral images are collected, some datasets may not be publicly accessible. Examples include data protected by privacy concerns and paid datasets. This would prevent models trained during the initial training phase from being reproduced, thereby hindering the transfer of knowledge. Consequently, it is essential to develop an incremental learning method that operates independently of old class samples. Furthermore, inspired by decoupled knowledge distillation, this paper introduces exclusion operations during the calculation of the distillation loss in the incremental learning process. These operations serve to eliminate misleading knowledge that could otherwise distort the loss computation, thereby enhancing the accuracy of distillation. In summary, the proposed method enables incremental learning to achieve more precise distillation loss calculation while eliminating reliance on existing class samples and mitigating catastrophic forgetting. The diagram below illustrates the framework of the primary algorithm employed in this study.

Figure 1. Main Algorithm Block Diagram

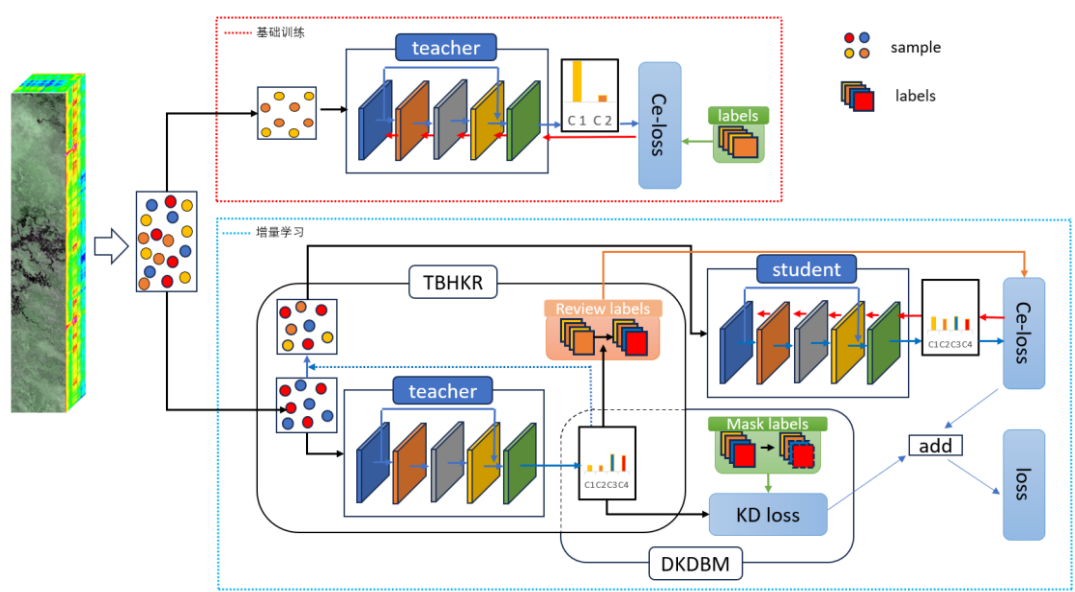

The algorithm is principally composed of two parts: the initial training phase and the incremental learning phase. The initial training phase corresponds to the training of the teacher model as seen in traditional knowledge distillation. The incremental learning phase employs two principal techniques: a method for retaining historical knowledge based on the teacher model and a masked partial class knowledge distillation algorithm. The complete training dataset is partitioned into two segments, each allocated to the respective phases. Following the completion of the initial training by the teacher network, the student network inherits the trained weights of the teacher and proceeds with training during the incremental learning phase. Given that the teacher network has acquired initial class knowledge during the initial phase, it inhibits the student network's tendency to forget this knowledge in the subsequent phase. Additionally, the masked method is used to eliminate incorrect outputs from the teacher model, thereby preventing interference with the training of the student model.

A. Teacher-Based Historical Knowledge Retention Method (TBHKR)

In incremental learning tasks involving new classes, the learning of incremental class samples during the incremental learning phase affects the knowledge acquired on the foundational class samples during the initial training phase. Specifically, in the absence of effective guidance for learning patterns pertinent to the incremental classes, the neural network inevitably experiences forgetting of previously learned information, a phenomenon known as catastrophic forgetting. To mitigate this issue, this paper proposes a teacher-based method for the retention of historical knowledge in class incremental learning. This method endows the neural network with modal information analogous to that of the base training classes during the incremental phase, even when the original base training samples are unavailable. Such a method helps the network maintain the retention of relevant knowledge. The primary schematic diagram of this method is illustrated below.

Figure 2. Main Block Diagram of the TBHKR Method

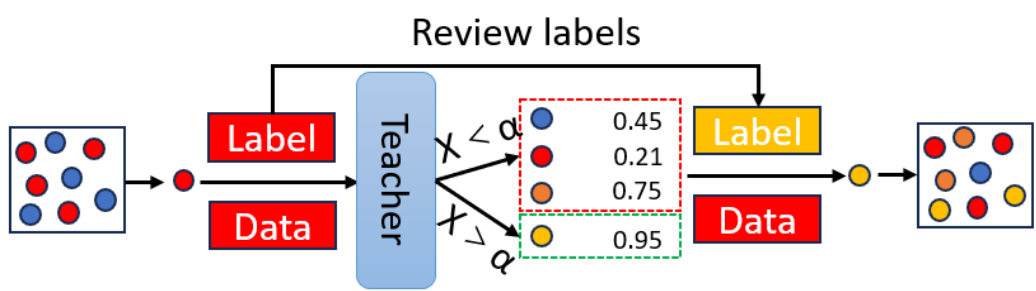

New class samples comprise data and labels. Upon inputting these samples into the teacher model, a set of output values, X, is generated. These outputs are compared against the threshold α. Samples with outputs exceeding α are retained, and their labels are subsequently modified to match the label of the most similar sample from the old class samples. The student model inherits the teacher model's weights and continues training using samples from the incremental learning phase. As these weights encompass knowledge pertaining to the initial class samples acquired by the teacher network during the baseline training phase, they are instrumental in evaluating modal information relevant to these classes. This process enables the student model to recognize samples in the new class that exhibit alignment with the modal characteristics of the old class samples. As demonstrated in the above figure, this method assists the student network in mitigating the forgetting of modal knowledge pertaining to the old classes. Specifically, the teacher network with fixed weights evaluates the similarity of new class samples to old classes. If a sample xi scores higher than a predetermined threshold a for class yi, it is regarded as having substantial modal information relevance to yi and may be utilized to supply analogous modal knowledge for that class. Consequently, the sample's class label is updated to yi for training purposes during the incremental learning phase. Throughout this training process, the student network not only assimilates modal knowledge regarding new classes but also acquires modal knowledge that is akin to that of previous classes. This process facilitates the associative recall of prior knowledge while acquiring new knowledge, thereby mitigating the effects of catastrophic forgetting.

B. Partial Class Knowledge Distillation Algorithm Based on Masking (DKDBM)

Based on the findings from Decoupled Knowledge Distillation [48], utilizing TCKD independently not only fails to produce advantageous effects but also has adverse impacts on student models. This paper hypothesizes that certain components necessitate decoupling during the incremental learning phase. Experimental results validate the presence of such components requiring decoupling, with post-decoupling accuracy exhibiting significant improvement. As exemplified by DKDBM in Figure 1, during incremental learning, the knowledge distillation (kd) loss is calculated in conjunction with the output of the teacher model and the masked labels. The labels masked by the mask labels correspond to those of the new class samples. Since the teacher model has not encountered new class samples during the initial stages of incremental learning, its outputs for these samples contain considerable inaccuracies. Consequently, masking the outputs for new class samples ensures that the computation of the distillation loss remains as precise as possible. Technical implementation details are provided below.

Initially, in knowledge distillation-based incremental learning for classification, let D represent the comprehensive set of classes throughout the process, P denote the set of classes during the baseline training phase, and N denote the set of classes during the incremental learning phase, where P ∪ N = D. Within set D, let $p_k$ represent the probability of belonging to set P, and $p_{\bar{k}}$ represent the probability of not belonging to set P, where k ∈ M. To facilitate formula analysis and emphasize key points, the subsequent introduction omits the distillation temperature from knowledge distillation. The formula is presented below.

$$p_k = \frac{\exp(z_k)}{\sum_{j \in D} \exp(z_j)} \tag{3-1}$$

$$p_{\bar{k}} = \frac{\sum_{i \in N} \exp(z_i)}{\sum_{j \in D} \exp(z_j)} \tag{3-2}$$

In set N, let $p_j$ be the probability of each class in set N, j∈N, as shown in the following formula.

$$p_j = \frac{\exp(z_j)}{\sum_{g \in N} \exp(z_g)} \tag{3-3}$$

Then, let $P^T$ be the probability output of the teacher model, and $P^S$ be the probability output of the student model. The distillation loss KL $(P^T||P^S)$ is given by the following formula.

$$\text{KL}\ (P^T||P^S) = \sum_{k \in M} p_k^t \log\left(\frac{p_k^t}{p_k^s}\right) + \sum_{i \in N} p_i^t \log\left(\frac{p_i^t}{p_i^s}\right) \tag{3-4}$$

By combining equations (3-1), (3-2), and (3-3), equation (3-4) can be decoupled to obtain equation (3-5), where $p_j = p_i/p_{\bar{k}}$ ,as shown below.

$$\text{KL}\ (P^T||P^S) = \sum_{k \in M} p_k^t \log\left(\frac{p_k^t}{p_k^s}\right) + p_{\bar{k}}^t \log\left(\frac{p_{\bar{k}}^t}{p_{\bar{k}}^s}\right) + p_{\bar{k}}^t \sum_{j \in N} p_j^t \log\left(\frac{p_j^t}{p_j^s}\right) \tag{3-5}$$

Equation (3-5) can be decomposed into three terms. Upon careful examination, it is readily apparent that the first term represents the KL divergence between the classes in the incremental learning phase and the entire set, the second term represents the KL divergence between the classes in the baseline training phase and the entire set, and the third term represents the KL divergence between the classes in the baseline training phase and the set

N. However, for the teacher model during the incremental learning phase, since it has never seen samples from the new class, its output information for the new class is incorrect. This can mislead the student model. The proposed masked partial class knowledge distillation algorithm can filter out this erroneous information, ensuring the correctness of the information transmitted by the teacher model.

4. Experiment

A. Datasets and Evaluation Metrics

This paper employs three datasets: Botswana, Houston, and Salinas. For classification tasks, the performance of the models is evaluated using three key metrics: overall accuracy (OA), average accuracy (AA), and the Kappa coefficient. The Kappa coefficient is specifically utilized to assess the reliability of the classifications.

B. Implementation Details

All experiments were conducted using PyTorch 2.6 and Python 3.11, with a system equipped with an NVIDIA GeForce RTX 4060 Laptop GPU, an AMD Ryzen 7 7840H CPU with Radeon 780M Graphics (8-core), and 32GB RAM. Data were sliced into 9×9 blocks and subjected to principal component analysis (PCA), yielding 20 principal components. During model training, the learning rates for the three datasets were set as follows: Salinas dataset at 1e-4, Houston and Botswana datasets at 5e-5. The number of training epochs was uniformly set to 100, with a batch size of 256 for all three datasets. This paper divided all classes within the dataset into two sets, designated for the baseline training phase and the incremental learning phase, respectively.

C. Comparative Experiment

To demonstrate the effectiveness of the proposed method, comparative experiments were conducted. Seven existing methods — LWF[16], EWC[49], iCaRL[24], PASS[50], DCPRN[41], End2End[51], and FORSTER[52] — were selected for comparison with the proposed method, totaling eight methods. Since the results for the first seven methods in the comparative experiments were averaged over five runs with their respective standard deviations, the proposed method was also evaluated over five runs to ensure comparable credibility. The proposed method comprises two stages: a baseline training stage and an incremental learning stage. To control variables, the classes selected for both stages matched those used in the comparative experiments. This comparative study employed the mean and variance of both OA and Kappa parameters to evaluate the overall model performance. Data points achieving the highest accuracy within a class were bolded and enlarged for easier observation. Details are illustrated in the figure below.

Table 1. Accuracy (%) of Different Methods for Each Class in the Botswana Dataset

|  | Class name | LWF | EWC | iCaRL | PASS | DCPRN | End2End | FOSTER | OURS |
|---|---|---|---|---|---|---|---|---|---|
| Base Classes | Reeds | 75.28+2.95 | 33.74+3.91 | 46.19+9.33 | 94.87±1.09 | 82.61±4.75 | 6.13+0.25 | 46.99+6.38 | **99.12±1.20** |
|  | Island interior | 41.34+2.19 | 24.06+2.17 | 31.90+15.53 | 6.21+10.77 | 77.17±2.67 | 4.51+0.42 | 78.13+8.09 | **99.70±0.67** |
|  | Acacia wood ands | 18.95+11.31 | 22.09+3.41 | 71.66+9.58 | 83.69+8.82 | 93.46±1.18 | 0.16+0.05 | 67.77+10.40 | **98.48±3.40** |
|  | Acacia grasslands | 12.26+11.37 | 31.93+3.86 | 41.64+8.17 | 72.92+37.30 | 67.54±3.90 | 2.44+1.17 | 83.54+10.27 | **99.94±0.13** |
|  | Exposed soils | 0.81+0.57 | 80.16+6.88 | 97.90+1.29 | 42.32+43.69 | 85.46±2.90 | 1.58+0.34 | 91.16+2.54 | **98.30±1.62** |
| Incremental Classes | Water | 3.47+4.05 | 87.50+4.33 | **99.54+0.16** | 98.81+0.86 | 84.61±4.52 | 34.70+2.31 | 75.04+3.75 | 96.80±4.69 |
|  | Hippo grass | 17.10±14.58 | 19.94+4.24 | 96.29+1.08 | 12.08+21.77 | 85.58±7.39 | 8.64+3.69 | 51.29+3.83 | **99.60±0.89** |
|  | Floodplain grasses1 | 11.80+12.11 | 28.93+1.03 | 93.23+0.94 | 78.73+16.19 | 52.36±6.65 | 25.74+2.49 | 66.69+25.40 | **99.52±0.66** |
|  | Floodplain grasses2 | 10.84+12.28 | 11.26+1.57 | **98.60±0.74** | 80.00+16.92 | 97.30±1.27 | 40.47+6.98 | 92.09+6.96 | 98.14±1.09 |
|  | Riparian | 29.78+16.70 | 31.46+0.29 | 59.20+12.63 | 54.57+11.03 | 78.15±3.76 | 3.20+1.27 | 40.60+6.41 | **98.00±1.98** |
|  | Firescar | 19.18+15.16 | 28.52+2.41 | 91.89+2.93 | 99.31+0.86 | 80.90±3.37 | 1.62+0.85 | 51.51+14.25 | **100.00±0.00** |
|  | Acacia shrublands | 2.90+1.63 | 17.64+2.11 | **99.60±0.01** | 92.50+2.81 | 74.70±6.34 | 2.81+0.09 | 83.39+8.91 | 98.64±2.39 |
|  | Short mopane | 7.37+3.43 | 35.98+3.06 | **98.07±1.49** | 82.21+15.19 | 76.08±2.43 | 3.83+0.28 | 96.24+2.69 | 96.56±1.30 |
|  | Mixed mopane | 13.16+1.38 | 29.76+1.74 | 81.44+15.79 | 93.88+4.43 | 65.55±8.26 | 1.87±0.76 | 61.94+9.77 | **95.90±2.89** |
| Performance | OA | 21.52+2.61 | 43.74+3.84 | 78.28+3.44 | 76.22+2.86 | 78.46±1.91 | 11.27+0.50 | 69.54±0.02 | **98.47±0.65** |
|  | Kappa | 14.99±0.03 | 39.18±0.04 | 76.32±0.04 | 74.19±3.10 | 77.58±0.02 | 5.69±0.00 | 67.03±0.02 | **98.34±0.70** |

Initially, during the initial training phase, the proposed model in this paper attained superior average accuracy rates across all five classes of the initial training phase in comparison to the other seven methods. Furthermore, for the Island interior and Acacia grasslands classes, the proposed method achieved accuracy enhancements exceeding 10% relative to the most effective of the previous seven methods, attaining 99.70 (+21.57) and 99.94 (+16.4), respectively. Additionally, the model demonstrated mean accuracy rates exceeding 98% across all initial training stages. The consistently high accuracy across the five classes indicates that the proposed model effectively mitigates catastrophic forgetting. During the incremental learning phase, the average accuracy of our method exceeded that of the other seven methods across all classes, with the exception of Water, Floodplain grasses2, Acacia shrublands, and Short mopane. Furthermore, for the Floodplain grasses2 class, our method attained an average accuracy of 98.14%, which is just under 0.5% below the top-performing method among the previous seven, iCaRL, which achieved an accuracy of 98.60%. For the classes Water, Acacia shrublands, and Short mopane, the average accuracy of the proposed method also exceeds 95%, with a gap of less than 3% compared to the best average accuracy among the first seven methods. For the Riparian class, the proposed method achieved a significant accuracy improvement. Among the seven methods, the best method, DCPRN, had an average accuracy of 78.15%, while the proposed method reached 98.00%, representing a direct increase of 19.85%—nearly one-fifth of the accuracy. Additionally, the variance in accuracy for this class was lower for the proposed method than for DCPRN, indicating greater stability. Finally, both the OA and Kappa parameters demonstrate the superiority of the proposed method over the other seven methods, with values of 98.47±0.65 and 98.34±0.70, respectively. Compared to the top seven methods, the mean OA score is 20.01 higher than the optimal method, and the mean Kappa score is 20.76 higher. Regarding variance, the standard deviation for both parameters remained below 1%. These experimental results confirm the stability of the proposed method.

Table 2. Accuracy (%) of Different Methods Across Classes in the Houston Dataset

|  | Class name | LWF | EWC | iCaRL | PASS | DCPRN | End2End | FOSTER | OURS |
|---|---|---|---|---|---|---|---|---|---|
| Base Classes | Trees | 37.30±7.40 | 31.75±4.49 | 33.14±10.41 | 81.45±2.98 | 52.36±6.65 | 9.93±12.35 | 81.03±1.58 | **98.22±3.06** |
|  | Soil | 72.30±5.29 | 44.23±3.71 | 57.33±14.05 | 66.02±18.96 | 97.30±1.27 | 8.77±8.32 | 77.58±4.26 | **99.90±0.22** |
|  | Water | 1.92±0.79 | 4.03±0.16 | 42.08±8.12 | 5.23±9.14 | 78.15±3.76 | 12.02±2.32 | 31.63±16.42 | **99.52±0.50** |
|  | Residential | 41.63±3.26 | 7.67±1.08 | 20.01±11.14 | 27.43±10.87 | 80.90±3.37 | 2.20±1.51 | 28.38±12.81 | **98.40±0.96** |
|  | Highway | 84.09±3.35 | 18.97±2.68 | 26.75±1.65 | 33.84±26.78 | 74.70±6.34 | 9.01±2.56 | 26.59±0.80 | **99.80±0.35** |
|  | Railway | 41.98±4.17 | 15.65±2.21 | 24.64±10.23 | 3.79±4.38 | 76.08±2.43 | 3.32±0.75 | 33.07±25.66 | **98.30±0.23** |
|  | Parking Lot 1 | 64.66±3.12 | 10.85±1.53 | 47.73±12.31 | 61.52±6.34 | 65.55±8.26 | 15.76±2.24 | 30.38±10.44 | **98.60±0.41** |
|  | Tennis Court | 22.02±8.13 | 10.73±1.75 | 84.81±2.82 | 28.46±29.81 | 94.06±1.55 | 15.22±1.93 | 83.88±5.79 | **99.76±0.54** |
| Incremental Classes | Healthy grass | 9.37±7.91 | 66.22±4.68 | 94.40±6.24 | 82.94±11.44 | 82.61±4.75 | 33.55±3.93 | 79.07±3.12 | **98.66±0.50** |
|  | Stressed grass | 15.37±5.33 | 28.97±3.96 | 85.55±4.12 | 88.28±7.44 | 77.17±2.67 | 1.50±1.93 | 42.94±24.89 | **94.04±4.05** |
|  | Synthetic grass | 1.29±0.96 | 40.11±3.21 | 98.32±0.78 | 97.42±2.37 | 93.46±1.18 | 3.49±6.58 | 92.86±1.76 | **99.98±0.04** |
|  | Commercial | 11.50±9.14 | 49.03±3.98 | 61.38±6.62 | 82.17±2.32 | 67.54±3.90 | 18.94±2.59 | 55.77±12.45 | **96.96±1.75** |
|  | Road | 2.92±3.11 | 15.38±2.17 | **93.73±1.91** | 85.91±3.04 | 85.46±2.90 | 4.68±5.12 | 32.24±14.81 | 82.12±4.04 |
|  | Parking Lot 2 | 11.99±1.32 | 23.72±3.35 | 84.81±2.83 | **86.99±5.77** | 84.61±4.52 | 3.63±2.86 | 54.20±14.87 | 86.82±2.50 |
|  | Running Track | 1.70±0.95 | 15.84±2.24 | 86.25±4.98 | 98.61±1.63 | 85.58±7.39 | 20.56±2.37 | 97.56±0.74 | **99.92±0.11** |
| Performance | OA | 34.18±0.93 | 28.08±1.51 | 56.69±2.77 | 63.35±3.86 | 78.46±1.91 | 13.80±0.86 | 54.06±2.94 | **96.66±0.63** |
|  | Kappa | 29.02±0.01 | 23.11±0.18 | 53.62±0.03 | 60.31±4.21 | 77.58±0.02 | 6.37±0.01 | 50.52±3.23 | **96.39±0.69** |

Initially, during the initial training stage, the model presented in this paper attained a higher average accuracy across all eight initial training classes in comparison with the other seven methods. Furthermore, for the Trees, Water, Residential, Highway, Railway, and Parking Lot 1 classes, the proposed method achieved accuracy improvements exceeding 10% relative to the most effective method among the prior seven methods, with respective gains of 98.22 (+17.19), 99.52 (+21.37), 98.40 (+17.50), 99.80 (+15.71), 98.30 (+22.22), and 98.60 (+33.05). Additionally, in terms of accuracy variance, aside from the Trees class, which exhibited a variance of 3.06, the variances for the other seven classes remained below 1%. This highlights the stability of the proposed method for class classification during the baseline training phase. Moreover, the mean accuracy during this phase surpassed 98%. The recognition accuracy across the five classes demonstrates the model's effectiveness in mitigating catastrophic forgetting. During the incremental learning phase, the mean accuracy for all classes except Road and Parking Lot 2 exceeded that of the other seven methods. Furthermore, for the Parking Lot 2 class, the mean accuracy achieved by our method was 86.82%, second only to the best method among the first seven, PASS, which recorded 86.99%. The difference between them was less than 0.5%. Additionally, the variance in accuracy for the proposed method's experiments was 2.5, which is less than half of the variance observed in the best method among the first seven (PASS), at 5.77. These findings demonstrate that the proposed method exhibits greater stability than PASS in recognizing the Parking Lot 2 class. Finally, in terms of the two parameters OA and Kappa, the proposed method outperformed the other seven methods, attaining values of 96.66±0.63 and 96.39±0.69, respectively. Compared to the previous seven methods, the mean OA score surpasses the optimal method by 18.20, and the mean Kappa score exceeds it by 18.81. Furthermore, the variance for both parameters remains below 1%, effectively demonstrating the robustness of the proposed method.

Table 3. Accuracy (%) of Different Methods Across Classes in the Salinas Dataset

| | Class name | LWF | EWC | iCaRL | PASS | DCPRN | End2End | FOSTER | OURS |
|---|---|---|---|---|---|---|---|---|---|
| Base Classes | Brocoli_green_weeds_1 | 63.66±3.80 | 75.00±4.30 | 68.59±1.78 | 1.44±2.89 | 95.01+6.15 | 7.10±4.04 | 67.65+39.32 | **99.94±0.09** |
| | Brocoli_green_weeds_2 | 0.54±0.22 | 34.10±3.75 | 56.49±1.25 | 35.73±43.77 | 77.17+2.67 | 6.19±6.35 | 41.25+33.12 | **99.94±0.09** |
| | Fallow_rough_plow | 59.54±2.57 | 54.73±1.49 | 86.76±1.90 | 57.66±37.54 | 93.10±3.57 | 9.97±0.99 | 95.01±2.78 | **98.40±0.16** |
| | Soil_vinyard_develop | 44.71±3.59 | 51.61±4.06 | 70.34 ±3.24 | 0.16 ±0.32 | 83.86+1.92 | 5.01±0.59 | 80.07 +31.39 | **99.94±0.09** |
| | Corn_senesced_green_weeds | 40.55±3.79 | 18.97±3.25 | 26.19±1.63 | 76.43±6.03 | 86.40 +1.12 | 3.87±1.30 | 76.58±5.38 | **99.88±0.13** |
| | Vinyard_untrained | 6.74±3.28 | 33.15±3.75 | 71.77 ±1.22 | 5.93 ±9.66 | **91.69±4.02** | 17.07 ± 2.41 | 89.03 +15.41 | 89.40±0.16 |
| | Vinyard_vertical_trellis | 16.14±0.78 | 22.14±2.84 | 43.41±1.74 | 24.22±30.01 | 58.58+1.96 | 4.98±0.69 | 56.70+23.77 | **99.94±0.09** |
| Incremental Classes | Fallow | 0.17±0.24 | 24.00±4.15 | 74.66±2.02 | 92.52±5.85 | 83.90+1.24 | 73.38±3.71 | 1.57+2.32 | **99.78±0.13** |
| | Fallow_smooth | 12.47±1.46 | 48.85±4.88 | 89.73 ±4.01 | 94.17±5.64 | 93.10±4.22 | 62.90±3.09 | 40.54+21.30 | **99.30±0.16** |
| | Stubble | 90.25±4.40 | 25.00±2.34 | 99.68±0.26 | 99.26±0.58 | 99.99+0.01 | 9.83±2.45 | 84.04±4.32 | **99.60±0.16** |
| | Celery | 85.59±8.23 | 24.94±4.31 | 97.31±0.72 | 98.82±2.11 | 99.77+0.40 | 10.54±6.42 | 82.84±4.42 | **99.96±0.05** |
| | Grapes_untrained | 87.34±3.56 | 47.42±4.71 | 92.78±2.15 | 85.22±8.02 | 95.51 +1.72 | 52.01±3.69 | 72.05+20.71 | **99.60±0.16** |
| | Lettuce_romaine_4wk | 0.53±0.24 | 39.19±1.24 | 67.07±3.10 | 93.05±2.93 | 42.66±3.35 | 52.35±3.94 | 0.13±0.17 | **99.30±0.16** |
| | Lettuce_romaine_5wk | 1.86±2.55 | 24.92±4.31 | 94.39±3.36 | **98.32±1.76** | 58.16±2.56 | 13.44 ± 1.23 | 81.42+14.28 | 97.80±0.16 |
| | Lettuce_romaine_6wk | 1.12±0.03 | 24.91±1.31 | 89.94±2.01 | **99.76±0.40** | 83.40+1.37 | 10.30±1.74 | 2.07+3.02 | 99.00±0.16 |
| | Lettuce_romaine_7wk | 10.90±4.43 | 25.76±3.18 | 95.71 ±1.10 | 60.61+30.71 | 80.92 +1.19 | 22.18±1.62 | 22.15+23.06 | **99.88±0.13** |
| Performance | OA | 45.31±2.28 | 41.93±3.00 | 83.70±1.02 | 62.83±6.27 | 85.40±2.03 | 15.21±2.62 | 58.83+7.49 | **98.27±0.16** |
| | Kappa | 37.26±0.03 | 33.09±0.04 | 82.41±0.01 | 58.93±7.07 | 84.47±0.02 | 18.98±0.13 | 53.86+7.49 | **98.08±0.16** |

Initially, during the foundational training phase, the model introduced in this paper achieved higher average accuracy than the other seven methods across all six classes, with the exception of Vinyard_untrained, furthermore, in the four classes — Brocoli_green_weeds_2, Soil_vinyard_develop, Corn_senesced_green_weeds, and Vinyard_vertical_trellis — the proposed method demonstrated accuracy improvements exceeding 10% relative to the best results obtained by the previous seven methods, with recorded values of 99.94 (+22.77), 99.94 (+16.08), 99.88 (+13.48), and 99.94 (+41.36), respectively. Concerning the Vinyard_untrained class, where performance was marginally inferior to the top seven methods, the proposed method achieved an average accuracy of 89.40. This figure is just under 2.5% below the highest accuracy among the top seven methods, which was DCPRN at 91.69. Additionally, the accuracy variance across all seven classes remained below 0.2%, thereby indicating the stability of the proposed method across classes within the foundational training phase. Consequently, the proposed model effectively alleviates catastrophic forgetting. During the incremental learning phase, the average accuracy across all classes, except for Lettuce_romaine_5wk and Lettuce_romaine_6wk, surpassed that of the other seven methods. Furthermore, for the classes Lettuce_romaine_5wk and Lettuce_romaine_6wk, our method achieved accuracies of 97.80% and 99.00%, respectively, which are less than 1% below the highest results obtained by the previous seven methods (PASS at 98.32% and 99.76%). Additionally, concerning accuracy variance, our method consistently demonstrates lower variance than the PASS method, indicating superior stability in experimental conditions. Lastly, regarding the Overall Accuracy (OA) and Kappa statistics, the proposed method outperforms the other seven methods with values of 98.27±0.16 and 98.08±0.16, respectively. In comparison to the top-performing methods, the mean OA score is 12.87 points higher, and the mean Kappa score is 13.61 points higher. Moreover, the variance for both parameters remains below 0.2%, clearly affirming the stability of the proposed method.

The experimental data for the first seven methods shown in the figure above are all taken from the publicly published paper[41]. Observing the performance of the proposed method on the three datasets mentioned above, it achieves overall higher gains than the first seven methods, both in the baseline training phase and the incremental learning phase.

D. Slice Size Experiment

Regarding data preparation, experimental results indicate differing performances across various slice sizes. To ascertain the most suitable slice size, comparative tests were conducted on multiple configurations. The results of these experiments are presented below, with the performance metric depicted in the figure as the Overall Accuracy (OA) percentage.

Figure 3. Model performance OA (%) for different slice sizes.

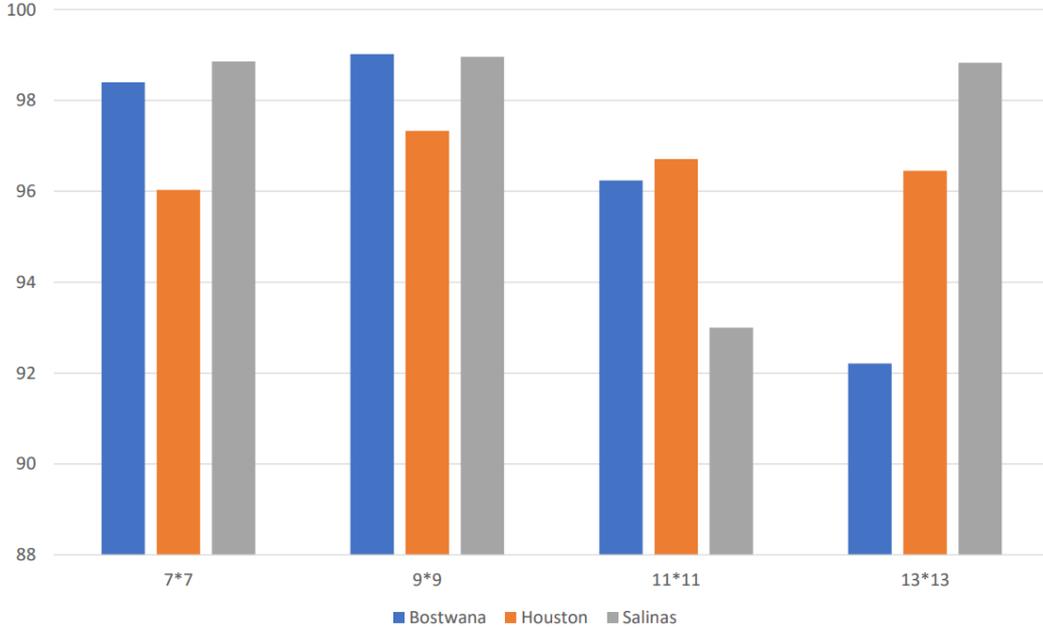

As illustrated in Figure 3, for the Botswana, Houston, and Salinas datasets, selecting a 9×9 slice size results in the highest overall accuracy (OA) across all three datasets. For the Botswana dataset, both 11×11 and 13×13 slice sizes produce OA values below 98%, showing a declining trend. The 7×7 tile size yields a lower OA value than the 9×9 tile size. In the Houston dataset, the 11×11 tile size markedly underperforms relative to 7×7, 9×9, and 13×13 tile sizes. Among the 7×7, 9×9, and 13×13 tile sizes, the 9×9 tile size still attains the highest OA value. Concerning the Salinas dataset, both 11×11 and 13×13 slice sizes achieved OA values below 97, demonstrating a descending trend. The 7×7 slice size recorded the lowest OA value among the four options. Experimental results indicate that a 9×9 slice size enables the model to attain optimal performance. Consequently, this paper employs a 9×9 slice size for model training across all three datasets.

E. Ablation Experiment

This paper presents ablation experiments conducted on the Salinas dataset, illustrating the contributions of various modules through the evaluation of three performance metrics: OA, AA, and Kappa. The results are depicted in the figure below.

Table 4. Ablation Experiments on the Salinas Dataset

| Dataset | Model | | | Accuracy | | |
|---------|-------|---|---|----------|---|---|
| | Baseline | Review lables | Mask lables | OA | AA | Kappa |
| Salinas | ✓ | | | 59.71±4.28 | 55.23±4.56 | 54.83±5.31 |
| | ✓ | ✓ | | 66.15±1.86 | 61.96±1.25 | 62.19±2.05 |
| | ✓ | | ✓ | 94.78±5.37 | 95.50±3.06 | 94.15±6.04 |
| | ✓ | ✓ | ✓ | 98.70±0.51 | 99.23±0.28 | 98.55±0.57 |

As illustrated in Table 4, experiments conducted without Review labels and Mask labels models serve as the baseline within this study. Four sets of experiments were performed, each repeated three times. The mean and variance were calculated to account for experimental variability while ensuring data validity. The method outlined in this paper comprises two primary components: Review labels and Mask labels. The experimental results depicted in the figure above were obtained from four independent runs. Upon introducing Review labels into the Salinas dataset, the mean values of OA, AA, and Kappa changed to 66.15 (+0.44), 61.96 (+6.73), and 62.19 (+7.36), respectively. Following the incorporation of Mask labels, the mean values of OA, AA, and Kappa shifted to 94.78 (+35.07), 95.50 (+40.27), and 94.15 (+39.32), respectively. When both models were integrated, the mean values of OA, AA, and Kappa adjusted to 98.70 (+38.99), 99.23 (+44), and 98.55 (+43.72), respectively. After introducing both models, the mean values of OA, AA, and Kappa changed to 98.70 (+38.99), 99.23 (+44), and 98.55 (+43.72), respectively. Overall, incorporating both Review Labels and Mask Labels into the three datasets yielded positive effects on the experiments, with the Mask Labels model contributing a higher accuracy boost. Notably, after incorporating both models, the mean accuracy of OA, AA, and Kappa metrics all exceeded 98%, while their variance narrowed to within 1%. The experimental results clearly demonstrate that the proposed model significantly enhances accuracy for hyperspectral image classification tasks while ensuring the stability of this high accuracy.

F. Conclusion

This paper proposes a mask-based partial classes knowledge distillation algorithm and a teacher-based knowledge retention method for incremental image classification. These methods effectively enhance model accuracy while addressing the model's dependency on old class samples. Specifically, the mask-based partial classes knowledge distillation algorithm enhances the student model's accuracy by filtering out potentially erroneous information during incremental learning. The teacher-based knowledge retention method for incremental image classification efficiently utilises new class samples without building a new data generation network, thereby resolving the model's dependency on existing class samples. The proposed methods achieved outstanding results across three datasets, demonstrating their feasibility.